\documentclass[a4paper,conference]{IEEEtran}

\ifCLASSINFOpdf
\else
\fi

\usepackage[english]{babel}
\usepackage{booktabs}
\usepackage{gensymb}
\usepackage{bm}
\usepackage{epsfig}
\usepackage{graphicx}
\usepackage{subcaption}
\usepackage{pgffor}
\usepackage{verbatim}

\usepackage{amsmath}

\usepackage{xspace}

\makeatletter
\DeclareRobustCommand\onedot{\futurelet\@let@token\@onedot}
\def\@onedot{\ifx\@let@token.\else.\null\fi\xspace}

\def\ie{{i.e}\onedot}

\def\etal{{et al}\onedot}
\makeatother

\makeatletter
\let\MYcaption\@makecaption
\makeatother

\begin{document}
%
% paper title
\title{Joint Voxel and Coordinate Regression for Accurate 3D Facial Landmark Localization}

% author names and affiliations
% use a multiple column layout for up to three different
% affiliations
\author{\IEEEauthorblockN{Hongwen Zhang\IEEEauthorrefmark{2}\IEEEauthorrefmark{3},
		Qi Li\IEEEauthorrefmark{2},
		Zhenan Sun\IEEEauthorrefmark{2}\IEEEauthorrefmark{3}\IEEEauthorrefmark{4}}
\IEEEauthorblockA{
	\IEEEauthorrefmark{2}Center for Research on Intelligent Perception and Computing (CRIPAC)\\
	\IEEEauthorrefmark{2}National Laboratory of Pattern Recognition (NLPR)\\
	\IEEEauthorrefmark{2}Institute of Automation, Chinese Academy of Sciences (CASIA)\\
	\IEEEauthorrefmark{3}University of Chinese Academy of Sciences (UCAS)\\
	\IEEEauthorrefmark{4}Center for Excellence in Brain Science and Intelligence Technology (CEBSIT), CAS\\
	Email: hongwen.zhang@cripac.ia.ac.cn, \{qli, znsun\}@nlpr.ia.ac.cn}	
}

% make the title area
\maketitle

% As a general rule, do not put math, special symbols or citations
% in the abstract
\begin{abstract}
3D face shape is more expressive and viewpoint-consistent than its 2D counterpart. However, 3D facial landmark localization in a single image is challenging due to the ambiguous nature of landmarks under 3D perspective. Existing approaches typically adopt a suboptimal two-step strategy, performing 2D landmark localization followed by depth estimation. In this paper, we propose the Joint Voxel and Coordinate Regression (JVCR) method for 3D facial landmark localization, addressing it more effectively in an end-to-end fashion. First, a compact volumetric representation is proposed to encode the per-voxel likelihood of positions being the 3D landmarks. The dimensionality of such a representation is fixed regardless of the number of target landmarks, so that the curse of dimensionality could be avoided. Then, a stacked hourglass network is adopted to estimate the volumetric representation from coarse to fine, followed by a 3D convolution network that takes the estimated volume as input and regresses 3D coordinates of the face shape. In this way, the 3D structural constraints between landmarks could be learned by the neural network in a more efficient manner. Moreover, the proposed pipeline enables end-to-end training and improves the robustness and accuracy of 3D facial landmark localization. The effectiveness of our approach is validated on the 3DFAW and AFLW2000-3D datasets. Experimental results show that the proposed method achieves state-of-the-art performance in comparison with existing methods.
\end{abstract}

\IEEEpeerreviewmaketitle

\section{Introduction}
% no \IEEEPARstart
Facial landmark localization has been extensively studied in the last decades and significant progress has been made on solving this problem.
Though impressive performance is achieved in 2D face alignment recently~\cite{bulat2016convolutional,Bulat2017HowFar}, 3D landmark localization in a single image remains challenging due to the ambiguous nature of landmarks under 3D perspective.

Currently, state-of-the-art approaches to facial landmark localization are dominated by regression based methods, demonstrating their effectiveness on addressing common issues such as occlusions, large variations of appearance on face images in the wild. 
Among them, cascaded regression methods~\cite{xiong2013supervised,cao2014face,ren2014face} attempt to learn the mapping from shape-index features to the landmark coordinates.
Though these methods could achieve highly accurate results for nearly frontal face images, their performances are barely satisfactory when it comes to novel view images.
On the other hand, heatmap regression based methods~\cite{bulat2016convolutional,Bulat2017HowFar} estimate the heatmap for each individual landmark instead. Such heatmap representation encodes the likelihood of positions being a specific landmark.
The heatmap regression strategy avoids the inefficient learning of the non-linear mapping from feature space to landmark positions, which has greatly facilitated landmark localization problems including face alignment~\cite{bulat2016convolutional,Bulat2017HowFar} and human pose estimation~\cite{bulat2016human,newell2016stacked}.
Though these methods could work well when the facial parts are visible, they might produce blurred heatmaps when there are invisible landmarks due to occlusions, making it unstable and error-prone to estimate landmark positions from those multi-mode heatmaps.

For 3D landmark localization, popular approaches employ a two-step strategy lifting the 2D estimation to 3D shape.
These methods~\cite{zhao2016fast,bulat2016two,gou2016shape} typically perform 2D landmark localization at first and then obtain the 3D face shape through depth estimation or 3D face model fitting.
Though such strategy is effective, it is suboptimal and sensitive to the result of 2D landmark localization.
In~\cite{pavlakos2017coarse}, Pavlakos \etal extend the 2D heatmap to 3D space and show that predicting the body joints in a discretized 3D space could be more effective for 3D human pose estimation. 
However, directly extending the 2D heatmap to its 3D version for each landmark is cumbersome and memory-demanding especially when the number of landmarks increases.

\begin{figure*}[t]
	\begin{center}
		\includegraphics[height=38mm]{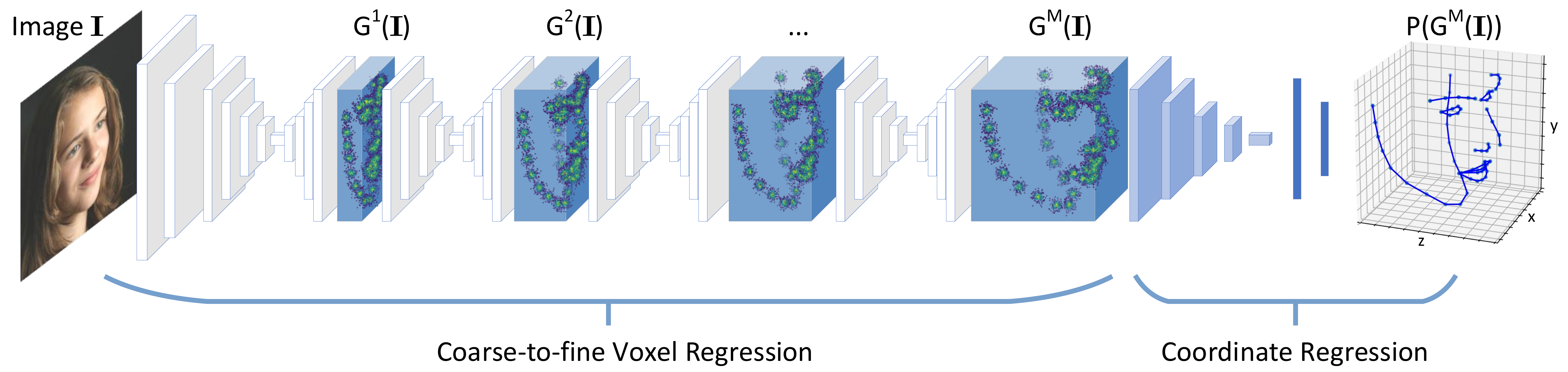}
		\caption{Pipeline of the Joint Voxel and Coordinate Regression (JVCR) method. The voxel regression subnetwork $G$ consists of $M$ Hourglass modules, which estimate the the compact volumetric representation from coarse to fine. The coordinate regression subnetwork $P$ takes as input the estimated volume and regresses 3D coordinates of the face shape.
		}
		\label{fig:framework}
	\end{center}
\end{figure*}

To cope with these limitations, we propose the Joint Voxel and Coordinate Regression (JVCR) method in this paper and make the following contributions towords accurate 3D facial landmark localization.
Firstly, we propose the compact volumetric representation which encodes the voxel-wise likelihood of positions being the target landmarks in 3D space.
The dimensionality of such a representation is fixed regardless of the number of landmarks, so that the required memory and computation could be reduced significantly.
Secondly, we adopt a coarse-to-fine strategy to regress the volumetric representation so that the 3D structural constraints between landmarks could be learned by the neural network more efficiently.
In addition, we employ 3D convolutions to regress the 3D coordinates of landmarks from the volumetric representation.
In this way, the 3D convolution network takes as input the entire volumetric representation of all landmarks, so that the prediction for those invisible landmarks could be more robust.
Finally, the proposed joint voxel and coordinate regression framework enables end-to-end training and shows promising results on 3D facial landmark localization.
Experimental results on 3DFAW~\cite{jeni2016first} and AFLW2000-3D~\cite{zhu2016face} datasets demonstrate the superiority of our approach.

The remainder of this paper is organized as follows. Section~\ref{RelatedWork} briefly reviews previous works related to ours.
The technical details of the proposed method are presented in Section~\ref{Method}.
Experimental results are reported in Section~\ref{Experiments}.
Finally, we conclude the paper in Section~\ref{Conclusion}.

\section{Related work}
\label{RelatedWork}

A significant amount of work has been introduced for landmark localization in the last decades.
In this section, we briefly review previous works related to ours, including methods for 2D and 3D landmark localization.

{\bf 2D landmark localization}. 
Typical methods for 2D face alignment include Constrained Local Models (CLMs)~\cite{cristinacce2006feature,saragih2011deformable,yu2013pose}, Active Appearance Models (AAMs)~\cite{cootes2001active,matthews2004active} and Cascaded Regression Methods (CRMs)~\cite{cao2014face,xiong2013supervised,ren2014face}. 
For CLMs and AAMs, they typically optimize the parametric representation of face shape iteratively according to the pre-trained appearance and shape models.
For CRMs, however, instead of representing the face shape parametricaly, they regress the facial landmark coordinates directly from shape-index features.
To avoid the inefficient learning of the pixel-to-coordinate mapping in CRMs, a majority of recent approaches~\cite{newell2016stacked,bulat2016convolutional,Bulat2017HowFar} cast landmark localization as regressing the heatmaps of landmarks instead of the coordinate vector.
Methods belonging to this type pursue regressing clear and accurate 2D heatmaps for the target landmarks.
For example, Stacked Hourglass Networks~\cite{newell2016stacked} uses the symmetric topology and intermediate supervision,
which has been demonstrated to be effective in both applications of human pose estimation~\cite{newell2016stacked} and face alignment~\cite{Bulat2017HowFar}.
Several state-of-the-art works~\cite{yang2017ed,Bulat2017HowFar,chu2017multi} built upon this architecture achieve nearly saturate performances on 2D landmark localization.
Despite their effectiveness, it is usually unstable to estimate the positions from multi-mode heatmaps for those invisible landmarks.
Very recently, instead of adopting the maximum operation, Sun \etal~\cite{sun2017integral} propose to infer the landmark coordinate from its heatmap through the integral operation, which allows end-to-end training and shows its effectiveness on 2D human pose estimation.

{\bf 3D landmark localization}.
One of the popular pipelines for 3D landmark localization adopts a two-stage strategy which performs the 2D landmark estimation at first and then predicts the depth information for these 2D landmarks.
In~\cite{zhao2016fast}, Zhao \etal propose a neural network to regress the 2D face shape firstly and then estimate the depth of the landmarks.
In~\cite{bulat2016two}, Bulat \etal introduce a two-stage method which performs the 2D heatmap regression followed by depth prediction.
Instead of estimating the depth information directly, Gou \etal~\cite{gou2016shape} propose to recover the 3D face shape by fitting the 3D morphable model to the 2D landmarks.
Moreover, cascaded regression methods are also extended to 3D landmark localization.
In~\cite{Tulyakov2017Viewpoint}, Tulyakov \etal propose the 3D shape invariant feature and estimate 3D face landmarks in a single step manner using the cascaded regressors.
On the other hand, Pavlakos \etal~\cite{pavlakos2017coarse} introduce the volumetric representation for 3D body joints and show that predicting the joints in a discretized 3D space could be more effective for 3D pose estimation.
The volumetric representation proposed in~\cite{pavlakos2017coarse} could be viewed as a natural extension of the 2D heatmap, which is highly demanding for memory and computation.
Though regressing such a representation in a coarse-to-fine manner could alleviate this problem~\cite{pavlakos2017coarse}, it still cannot avoid the curse of dimensionality when the number of target landmarks increases.
Hence it can not be easily generalized to other 3D object alignment problems.
Alternatively, we propose to encode the positions of all landmarks in a single volume with the dimensionality fixed regardless of the number of landmarks, providing a much more efficient solution for general 3D landmark localization.

\section{Method}
\label{Method}

In this section, we present the Joint Voxel and Coordinate Regression (JVCR) method for 3D facial landmark localization in detail.
The pipeline of the proposed method is shown in Fig.~\ref{fig:framework}.
Our network architecture consists of two main blocks, a voxel regression subnetwork that estimates the compact volumetric representation from coarse to fine, followed by a 3D convolution network that takes the estimated volume as input and regresses the coordinate vector.
In the following subsection, we begin with the description of the compact volumetric representation.
Then we introduce the voxel regression subnetwork, the coordinate regression subnetwork as well as the training scheme for the proposed method.

\begin{figure}[t]
	\centering
	\begin{subfigure}[b]{0.2\textwidth}
		\centering
		\includegraphics[width=1.1\textwidth]{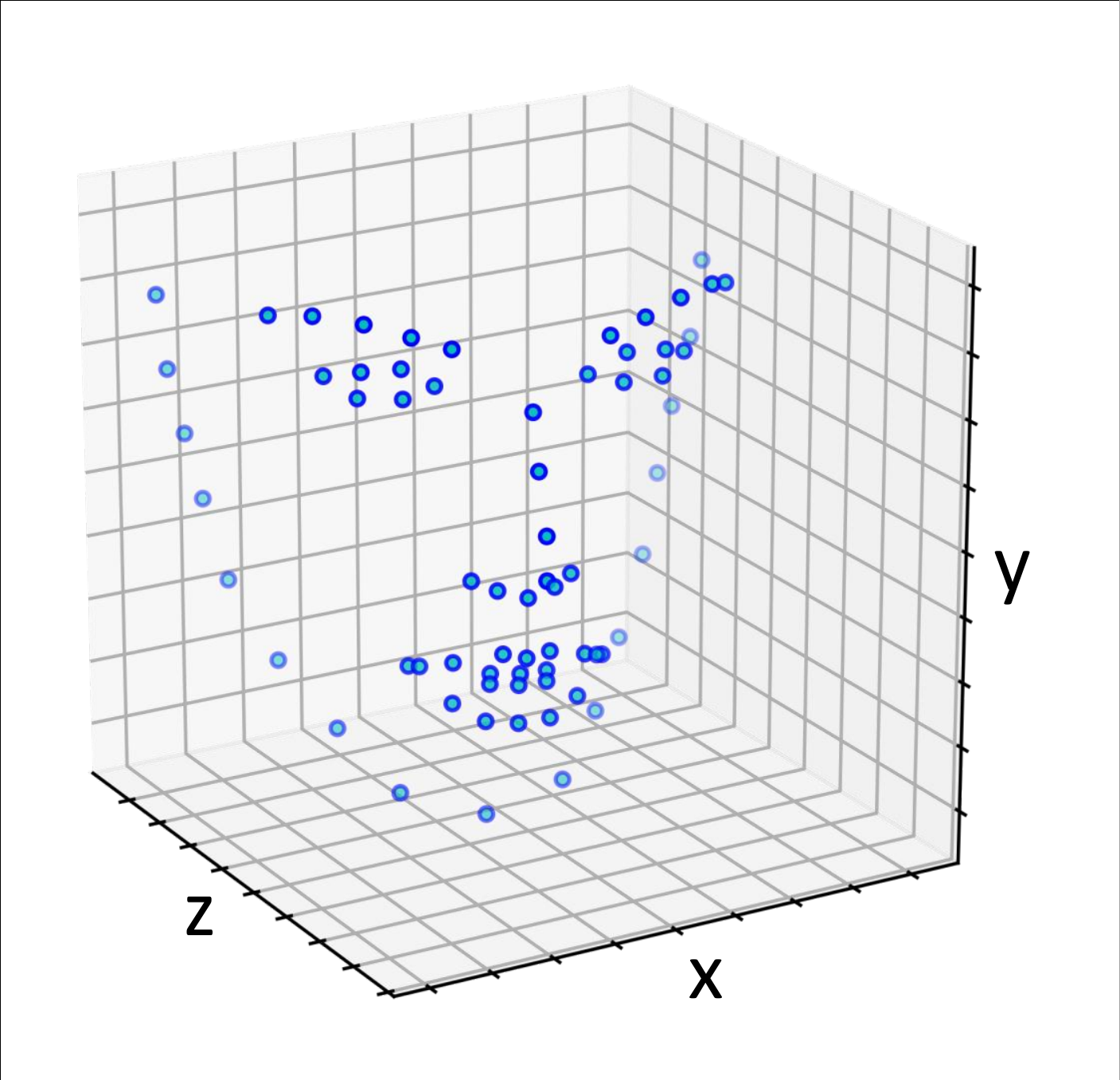}
		\caption{ }
		\label{fig:coordinate}
	\end{subfigure}
	\hspace{5mm}
	\begin{subfigure}[b]{0.2\textwidth}
		\centering
		\includegraphics[width=1.1\textwidth]{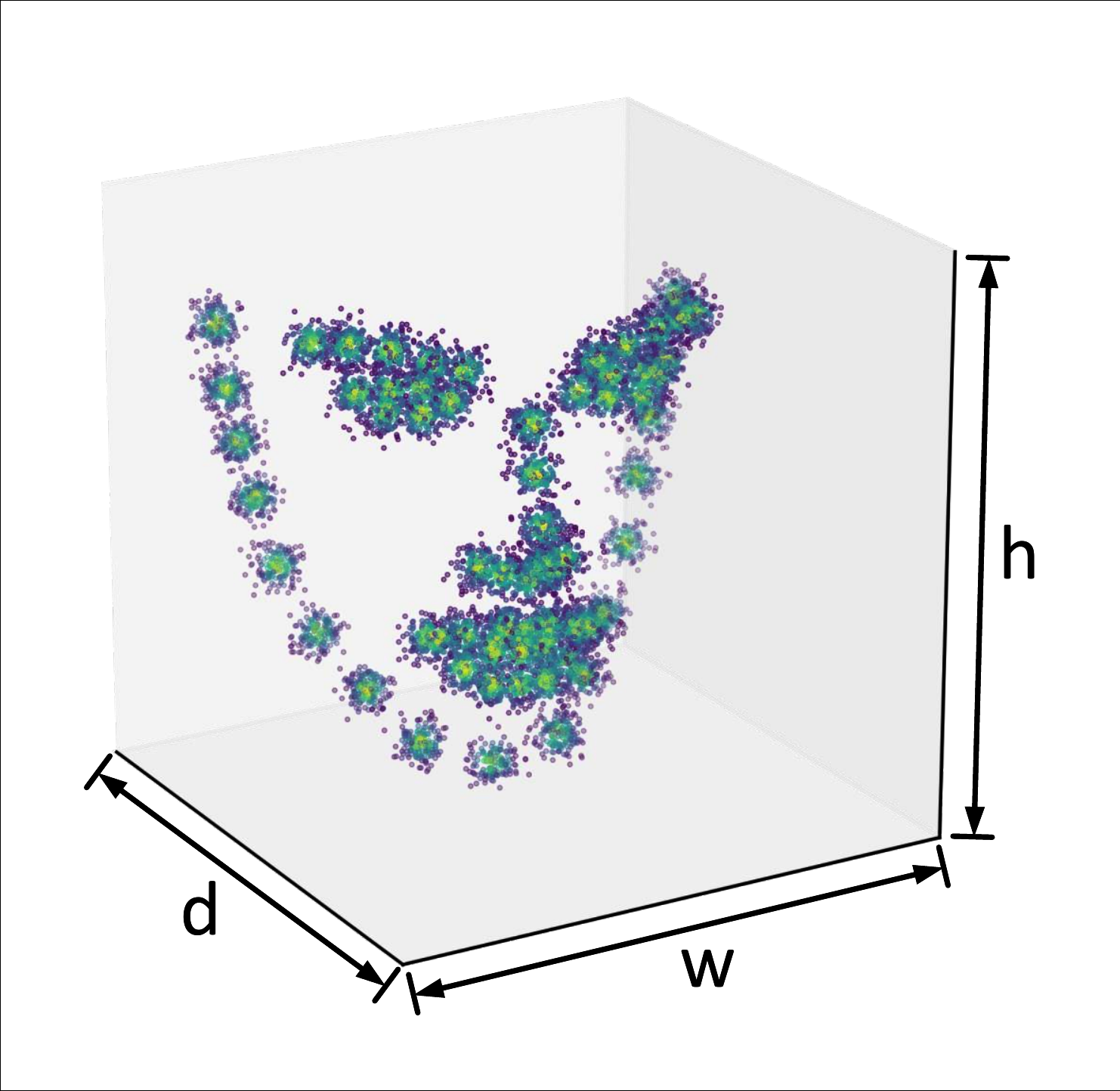}
		\caption{ }
		\label{fig:volumetric}
	\end{subfigure}
	\caption{Visualization of the 3D face shape (a) and the compact volumetric representation (b). The voxel values are indicated by the density of the point cloud. Best viewed in color.}
	\label{fig:visualVoxel}
\end{figure}

\subsection{Compact Volumetric Representation for 3D Face Shape}

Previous works~\cite{tompson2014joint,pavlakos2017coarse} have shown that encoding the landmark positions into the heatmap-like or volumetric representation could provide much more discriminative information than naively concatenating the coordinate vectors of 2D or 3D landmarks.
Such forms of supervision make it easier for fully convolutional networks to learn the pixel to pixel mapping,
and have been used in the context of both facial landmark localization~\cite{bulat2016convolutional} and human pose estimation~\cite{tompson2014joint}.

For 3D landmark localization, the volumetric representation proposed in~\cite{pavlakos2017coarse} encodes the position of a specific landmark in a volume with a 3D Gaussian centered around the groundtruth position. 
Though this idea extends the typically used 2D heatmap in a natural manner, it leads to a representation with large dimensionality.
Instead of representing each landmark individually, we propose a volumetric representation encoding the positions of all target landmarks in a more compact manner.
Specifically, coordinates of all the target landmarks are converted into a discretized 3D volume $\mathbf{V}$ with the size of $w\times h\times d$. 
Let $v_{i,j,k}$ denote the value of voxel $(i,j,k)$.
For the $n$-th landmark located at $\bm{x}_{gt}^{n}=(x,y,z)$, its contribution to $v_{i,j,k}$ can be written as:
\begin{equation}
\label{eq:voxel}
{v}_{i,j,k}^{n}=\frac{1}{2\pi {{\sigma }^{2}}}{{e}^{-\frac{{{(x-i)}^{2}}+{{(y-j)}^{2}}+{{(z-k)}^{2}}}{2{{\sigma }^{2}}}}}
\end{equation}
where the kernel size $\sigma$ could be set empirically.
For 3D face shape with $N$ target landmarks, the overall contribution to $v_{i,j,k}$ takes as the maximum value in $\{{v}_{i,j,k}^{n}\}_{n=1}^N$:
\begin{equation}
{v}_{i,j,k}=\underset{n}{\mathop{\max }}\,v_{i,j,k}^{n}
\end{equation}
In this way, the dimensionality of the representation is fixed regardless of the number of target landmarks. 
Fig.~\ref{fig:visualVoxel} visualizes the 3D face shape and the corresponding compact volumetric representation.
It is worth noting that, in this paper, we also refer to the compact volumetric representation as volumetric representation or volume for simplicity.

\subsection{Joint Voxel and Coordinate Regression}

Cascaded regression in a coarse-to-fine manner is wildly employed in 2D landmark localization.
Such a strategy could make full use of the regressors and progressively refine the output of the networks.
Our method follows this technique and decouples the 3D facial landmark localization problem into the following two sub-tasks.
The first one aims to regress the ideal volumetric representation of 3D landmarks in a coarse-to-fine manner.
The second one aims to regress the coordinates of landmarks from the volumetric representation.

\subsubsection{Coarse-to-fine Voxel Regression}
The voxel regression subnetwork $G$ learns the mapping from pixels of the face image $\mathbf{I}$ to the volumetric representation $\mathbf{V}$: 
$G(\mathbf{I})\rightarrow \mathbf{V}$.
Inspired by previous works~\cite{newell2016stacked,pavlakos2017coarse} on 2D and 3D human pose estimation, we also adopt the stacked hourglass networks~\cite{newell2016stacked} with intermediate supervision and skip connection.
Specifically, the voxel regression subnetwork consists of $M$ stacked Hourglass modules~\cite{newell2016stacked} of which supervisions are volumes denoted as $\{\mathbf{V}^m\}_{m=1}^M$.
Then, the voxel regression subnetwork is trained using the voxel-wise mean squared error loss:
\begin{equation}
\mathcal{L}_{vox} =\sum\limits_{m}^{M}{\sum\limits_{i,j,k}{{{\left\| {{G}^{m}}{{(\mathbf{I})}_{i,j,k}}-\mathbf{V}^{m}_{i,j,k} \right\|}^{2}}}}
\end{equation}
where $G^m(\cdot)$ denotes the volume outputted by the $m$-th Hourglass module. Noted that $G^M(\cdot)=G(\cdot)$ is equivalent to the final output of the voxel regression subnetwork.

As pointed out in~\cite{pavlakos2017coarse}, the prediction along the $z$ dimension is much more challenging than another two dimensions.
Hence, coarse-to-fine regressing the volumes with the increasing resolution along $z$ dimension could be more effective and robust.
In practice, the resolution $d$ of $\mathbf{V}^m$ takes number from preset values and progressively increases along with $m$.

\subsubsection{Coordinate Regression}

Typical heatmap regression based methods~\cite{bulat2016convolutional,wei2016convolutional,newell2016stacked} retrieve the coordinates of landmarks directly from the peak points of the corresponding heatmaps.
Considering that the positions of all landmarks are encoded into a single volume, the conventional ``tacking-maximum'' operation is no longer applicable in our case since the order of landmarks is not preserved in our compact volumetric representation. 
Hence, it needs to infer the coordinates of landmarks from the corresponding volumetric representation.

To this end, we propose a coordinate regression subnetwork $P$ to learn the mapping from the compact volumetric representation $\mathbf{V}$ to the corresponding coordinate vector $\bm{x}$:
$P(\mathbf{V})\rightarrow \bm{x}$.
Inspired by the work on 3D object recognition~\cite{wu20153d} and hand pose estimation~\cite{ge20173d}, we adopt the 3D convolution kernel instead of 2D convolution in our coordinate regression subnetwork.
The 3D convolution is typically used to extract features from both spatial and temporal dimensions for video analysis problems~\cite{ji20133d}.
Hence, the 3D convolution could be more naturally adopted to extract the 3D information from the volumetric representation.
The proposed coordinate regression subnetwork consists of five 3D convolution layers, with batch normalization and Leaky ReLU activation added in between and a fully connected layer at the end.
For training, we employ the $\mathcal{L}_2$ regression loss on the predicted coordinate vector:
\begin{equation}
{\mathcal{L}_{coord}}=\left\| {{\bm{x}}_{gt}}-P(\mathbf{V}) \right\|_{2}^{2}
\end{equation}
where $\bm{x}_{gt}$ denotes the concatenated vector of the ground-truth 3D landmark coordinates.

\subsubsection{Training}
Instead of training the whole network from scratch, we adopt a two-stage training scheme which is more stable and effective in our experiments.
The two subnetworks mentioned above are pre-trained separately for each sub-task beforehand and fine-tuned as an integrated one finally.
Specifically, at the pre-training stage, the voxel regression subnetwork is trained with the face images and the ground-truth volumes.
Meanwhile, the coordinate regression subnetwork is trained with the ground-truth volumes and the corresponding coordinate vectors.
At the fine-tuning stage, the coordinate regression subnetwork is attached to the voxel regression subnetwork, and the whole network is fine-tuned with the joint supervision of both the ground-truth volumes and coordinate vectors.
Formally, the whole network is trained in an end-to-end manner using the following loss function at the final stage:
\begin{equation}
\begin{split}
\mathcal{L}
&= \mathcal{L}_{vox} + \lambda \mathcal{L}_{coord} \\
&= \sum\limits_{m}^{M}{{{{\left\| {{G}^{m}}{{(\mathbf{I})}}-\mathbf{V}^{m} \right\|}_{2}^{2}}}} + \lambda \left\| {{\bm{x}}_{gt}}-P(G^M(\mathbf{I})) \right\|_{2}^{2}
\end{split}
\end{equation}
where $\lambda$ is used to balance the two terms.

\section{Experiments}
\label{Experiments}

In this section, the implementation detail of the proposed method are described firstly.
Then, the datasets as well as evaluation metrics used in our experiments are introduced.
Finally, we present the experimental results of the proposed method.

\subsection{Implementation Detail}
The proposed network takes as input a $256\times 256$ face image and outputs predictions of the volumetric representation and the coordinate vector.
Inspired by the setting of~\cite{newell2016stacked}, four Hourglass modules are stacked together as the voxel regression subnetwork (\ie $M=4$) and output volumes with the size of $64\times 64\times d$, where the resolution $d$ of $z$-dimension is chosen from the set $\{1,2,4,64\}$ successively. 
The Gaussian kernel size in Eq.~\ref{eq:voxel} is set to $\sigma =1$ in our experiments.
During training, data augmentation, such as rotation, scaling and fliping, was applied randomly to input images.
The network was traind for 25 epochs in total, including 15 epochs for the pre-training stage and 10 epochs for the fine-tuning stage, respectively.
We adopted the RMSprop~\cite{tieleman2012lecture} optimization algorithm with an initial learning rate of $2.5\times 10^{-4}$, which was reduced by a factor of 10 every 10 epochs.
Our approach was implemented using PyTorch. During testing, it takes about 50ms for our model to process an image on a TITAN Xp GPU.
Code is made publicly available\footnote{https://github.com/HongwenZhang/JVCR-3Dlandmark}.

\subsection{Datasets}

\textbf{3DFAW}~\cite{jeni2016first}. The 3DFAW dataset is provided by the 3D Face Alignment in the Wild (3DFAW) Challenge~\cite{jeni2016first} organizers, containing more than 23000 face images from BU-4DFE~\cite{yin2008high}, BP4D-Spontaneous~\cite{zhang2014bp4d} and MultiPIE~\cite{gross2010multi}. 
66 3D facial landmarks as well as the face bounding boxes are annotated for each face image.
The 3D points are annotated consistently using a model-based structure-from-motion technique~\cite{jeni2017dense}.
The 3DFAW dataset is divided into three subsets: the training set, the validation set and the test set, containing 13969, 4725 and 4912 face images, respectively.
Our method is trained on the training set and tested on both the validation and test set. 
It should be noted that the ground-truth 3D landmarks of the test set are not publicly available.
Hence the evaluation results on the test set are provided by 3DFAW Challenge organizers via the CodaLab platform\footnote{https://competitions.codalab.org/competitions/10261}.

\textbf{300W-LP}~\cite{zhu2016face}. The 300W-LP dataset contains 61225 synthesized face images across large poses ranging from $-$90\degree~to 90\degree. 
Those images are synthesized from 300W~\cite{sagonas2013300} using the 3D morphable model based profiling algorithm proposed in~\cite{zhu2016face}.
For each face, 68 3D landmarks are retrieved from the parameters of the 3D morphable model, using the released code of~\cite{zhu2016face}.
In our experiments, the depth values are normalized to have zero mean.
We only use this dataset for training and test our method on the AFLW2000-3D dataset mentioned bellow.

\textbf{AFLW2000-3D}~\cite{zhu2016face}. The AFLW2000-3D dataset contains 2000 face samples selected from the AFLW~\cite{kostinger2011annotated} dataset, introduced by Zhu \etal~\cite{zhu2016face} along with the 300W-LP dataset.
The 68 3D landmarks annotated in AFLW2000-3D are consistent with those of 300W-LP.
We use the AFLW2000-3D dataset only for testing in our experiments, following the common protocol in the literature~\cite{zhu2016face,Bhagavatula2017Faster}.

\subsection{Evaluation Metrics}
For fair comparison, the evaluation metrics are adopted in consistency with previous works~\cite{jeni2016first,zhu2016face}.
For 3D landmark localization, the Ground Truth Error (GTE) and Cross View Ground Truth Consistency Error (CVGTCE) are used to measure the performance as recommenced in the 3DFAW Challenge~\cite{jeni2016first}. 
The GTE is defined as the average point-to-point Euclidean error normalized by the distance between the outer corners of the eyes.
The CVGTCE is proposed in the 3DFAW Challenge and aims at evaluating the cross-view consistency of the predicted landmarks.
For evaluating the 2D projection of the 3D landmarks, the metric is the Normalized Mean Error (NME), which is defined as the average 2D point-to-point Euclidean error normalized by the square root of the bounding box size.

\begin{table}[h]
	\centering
	\caption{Comparison of GTE on the 3DFAW validation set. }
	\begin{tabular}{lc}
		\toprule
		Method & GTE (\%) \\
		\midrule
		SDM+3DMM~\cite{gou2016shape} & 6.34 \\
		Gou \etal~\cite{gou2016shape} & 5.90 \\
		Bulat \etal~\cite{bulat2016two} & 4.94 \\
		\midrule
		JVCR  & \bf 4.36 \\
		\bottomrule
	\end{tabular}%
	\label{tab:cmp3DFAWvalid}%
\end{table}%

\begin{table}[h]
	\centering
	\caption{Comparisons of CVGTCE and GTE on the 3DFAW test set.}
	\begin{tabular}{lcc}
		\toprule
		Method & CVGTCE (\%) & GTE (\%) \\
		\midrule
		Zavan \etal~\cite{de20163d} & 5.90 & 10.80 \\
		Gou \etal~\cite{gou2016shape} & 4.94 & 6.20 \\
		Zhao \etal~\cite{zhao2016fast} & 3.97  & 5.88 \\
		Bulat \etal~\cite{bulat2016two} & 3.47 & 4.56 \\
		Tulyakov \etal~\cite{Tulyakov2017Viewpoint} & 3.80   & 5.10\\
		\midrule
		JVCR  & \textbf{3.46}  & \textbf{4.35} \\
		\bottomrule
	\end{tabular}%
	\label{tab:cmp3DFAWtest}%
\end{table}%

\subsection{Experimental Results}
In this subsection, we compare our approach with existing methods including top ranked methods on the 3DFAW Challenge, state-of-the-art 2D face alignment method FAN~\cite{Bulat2017HowFar} and 3D face model based methods such as 3DDFA~\cite{zhu2016face} and 3DSTN~\cite{Bhagavatula2017Faster}.

\subsubsection{Evaluation on 3DFAW}
The evaluation on 3DFAW consists of two parts. 
The first part is evaluated on the validation set. 
The second part is evaluated on the test set, of which performance is provided by the challenge organizers.

Table~\ref{tab:cmp3DFAWvalid} shows the comparison results of GTE on the validation set.
It can be observed that the proposed method outperforms others, all of which are based on the two-step strategy. 
For comparison with top ranked methods on the 3DFAW Challenge, we further evaluate our method on the test set.
Comparisons of both the CVGTCE and GTE on the 3DFAW test set are reported in Table~\ref{tab:cmp3DFAWtest}.
Note that the ground truth 3D landmarks of the test set are not available to the participants, and the numbers for all methods are taken from the CodaLab leaderboard and the literature~\cite{jeni2016first,Tulyakov2017Viewpoint}.
As shown in Table~\ref{tab:cmp3DFAWtest}, the proposed method achieves the best result in comparison with other methods, including the previously top-ranked method~\cite{bulat2016two} and Tulyakov \etal~\cite{Tulyakov2017Viewpoint} which is built upon a 3D variant of cascaded regression method.
It demonstrates the effectiveness of the proposed JVCR framework for accurate 3D facial landmark localization.
Fig.~\ref{fig:3dfawDemo} shows example results of the proposed method on the 3DFAW test set.

\begin{table}[b]
	\centering
	\caption{Comparison of GTE on the AFLW2000-3D dataset. }
	\begin{tabular}{lc}
		\toprule
		Method & GTE (\%) \\
		\midrule
		FAN+Depth~\cite{Bulat2017HowFar} & 7.45 \\
		\midrule
		JVCR  & \bf 7.28 \\
		\bottomrule
	\end{tabular}%
	\label{tab:compAFLW3D}%
\end{table}%

\begin{figure}[t]
	\begin{center}
		\includegraphics[height=50mm]{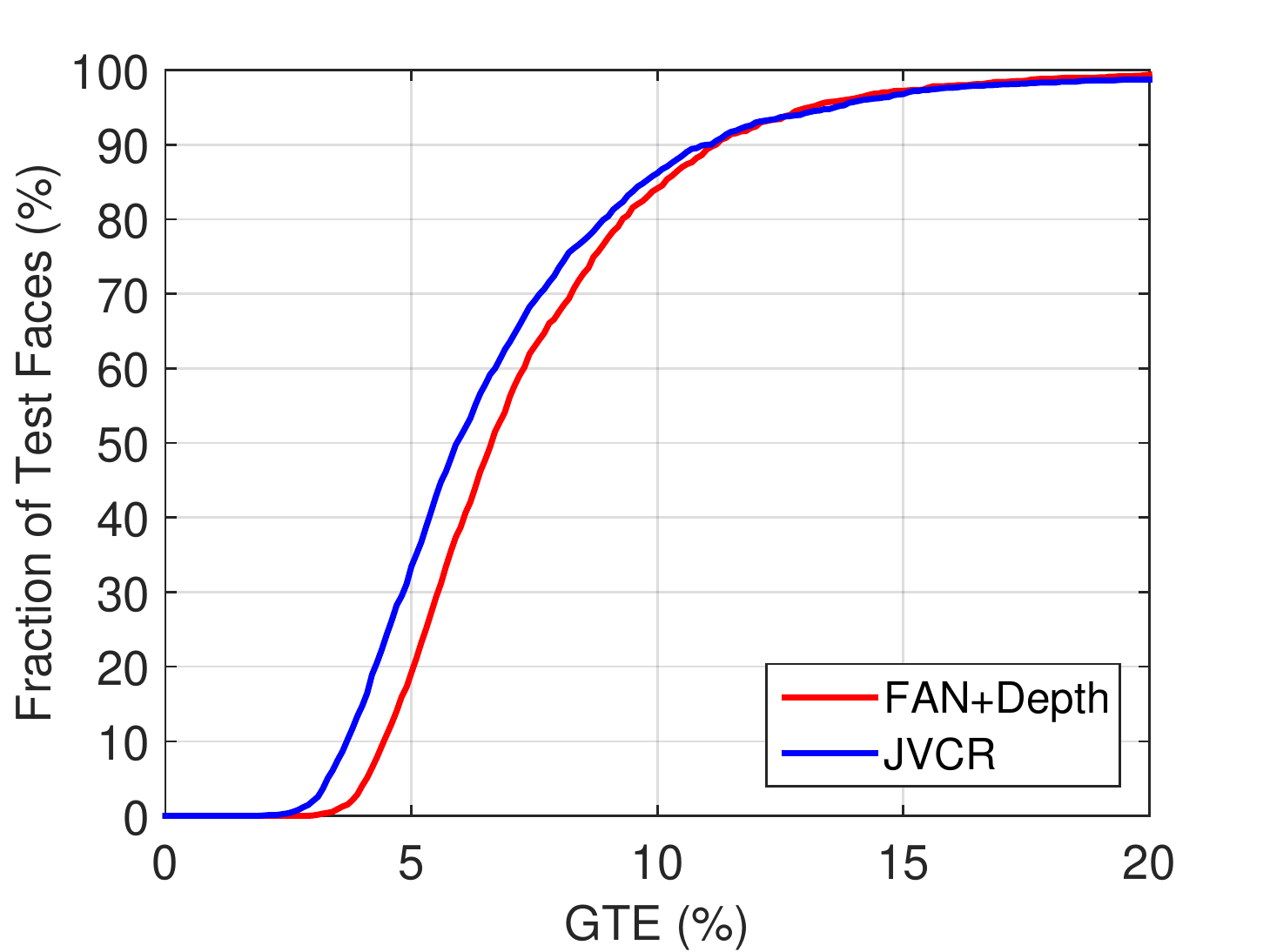}
		\caption{Comparison of cumulative errors distribution (CED) curves on AFLW2000-3D.
		}
		\label{fig:compCED}
	\end{center}
\end{figure}

\begin{table}[t]
	\centering
	\caption{Comparison of NME (\%) on the AFLW2000-3D dataset. Only 2D coordinates are involved in the evaluation.}
	\begin{tabular}{l|ccc|cc}
		\toprule
		& \multicolumn{5}{c}{AFLW2000-3D (68 pts)} \\
		\cmidrule{2-6}    Method & [0\degree,30\degree] & [30\degree,60\degree] & [60\degree,90\degree] & mean  & std \\
		\midrule
		RCPR~\cite{burgos2013robust}  & 4.26  & 5.96  & 13.18 & 7.80  & 4.74 \\
		ESR~\cite{cao2014face}   & 4.60  & 6.70  & 12.67 & 7.99  & 4.19 \\
		SDM~\cite{xiong2013supervised}   & 3.67  & 4.94  & 9.76  & 6.12  & 3.21 \\
		3DDFA~\cite{zhu2016face} & 3.78  & 4.54  & 7.93  & 5.42  & 2.21 \\
		3DDFA+SDM~\cite{zhu2016face} & 3.43  & 4.24  & 7.17  & 4.94  & 1.97 \\
		3DSTN~\cite{Bhagavatula2017Faster} & 3.15  & 4.33  & 5.98  & 4.49  & 1.42 \\
		FAN~\cite{Bulat2017HowFar} & \bf 2.77  & 3.48  & 4.60   & \bf 3.62  & 0.92 \\
		% DenseFA & -     & -     & -     & 4.5   & - \\
		\midrule
		JVCR  &  2.94  & \bf 3.46  & \bf 4.53  & 3.64  & \bf 0.81 \\
		\bottomrule
	\end{tabular}%
	\label{tab:compAFLW2D}%
\end{table}%

\iffalse
\begin{figure*}[t]
	\centering
	\foreach \idx in {1,2,3,4,5,6,7,8} {
		\begin{subfigure}[h]{0.11\textwidth}
			\centering
			\includegraphics[width=1.1\textwidth]{Img/demo/3dfaw/3dfaw_img\idx.pdf}
			\includegraphics[width=1.1\textwidth]{Img/demo/3dfaw/3dfaw_vox\idx.pdf}
			\includegraphics[width=1.1\textwidth]{Img/demo/3dfaw/3dfaw_coord\idx.pdf}
		\end{subfigure}
	}
	\caption{Example results of the proposed method on the 3DFAW dataset. The top row shows the face images as well as the 2D facial landmark localization results. The middle row shows the estimated volumetric representations. The last row shows the predicted 3D facial landmarks.}
	\label{fig:3dfawDemo}
\end{figure*}

\begin{figure*}[t]
	\centering
	\foreach \idx in {1,2,3,4,5,6,7,8} {
		\begin{subfigure}[h]{0.11\textwidth}
			\centering
			\includegraphics[width=1.1\textwidth]{Img/demo/aflw/aflw_img\idx.pdf}
			\includegraphics[width=1.1\textwidth]{Img/demo/aflw/aflw_vox\idx.pdf}
			\includegraphics[width=1.1\textwidth]{Img/demo/aflw/aflw_coord\idx.pdf}
		\end{subfigure}
	}
	\caption{Example results of the proposed method on the AFLW2000-3D dataset. Images are arranged as the same as Fig.~\ref{fig:3dfawDemo}}
	\label{fig:aflwDemo}
\end{figure*}
\fi

\begin{figure*}[t]
	\centering
	\includegraphics[width=0.99\textwidth]{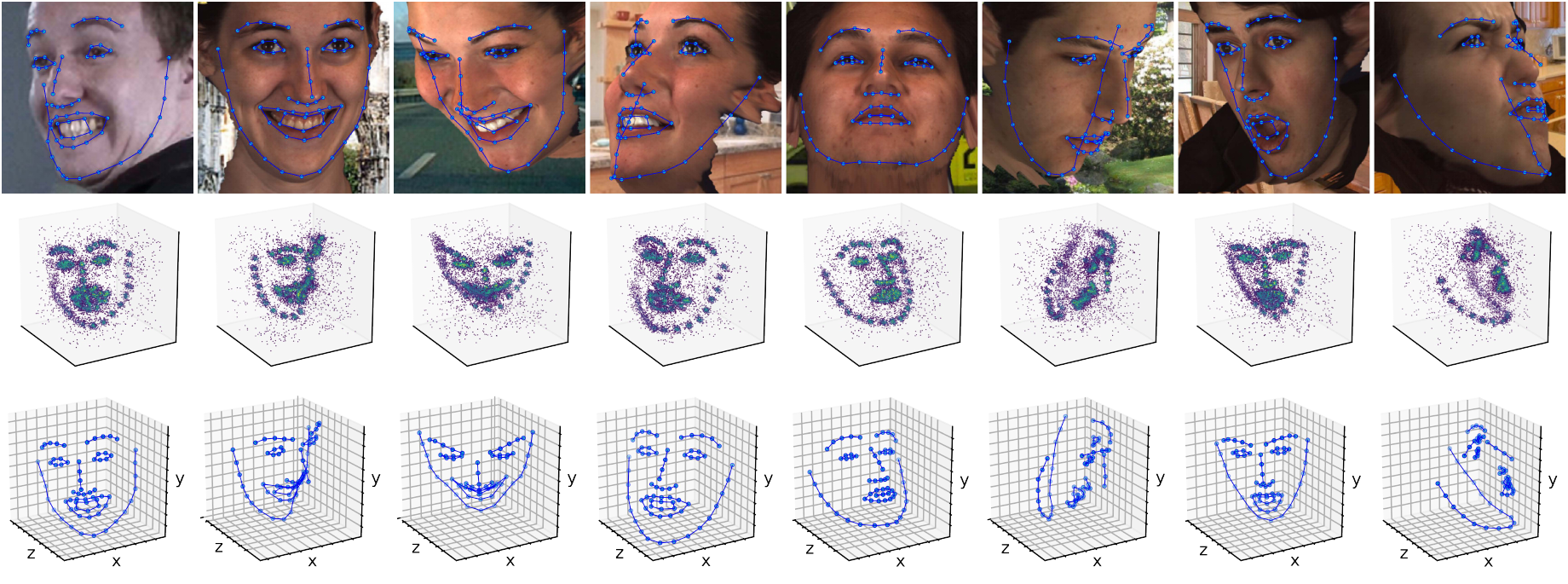}
	\caption{Example results of the proposed method on the 3DFAW dataset. The top row shows the face images as well as the 2D facial landmark localization results. The middle row shows the estimated volumetric representations. The last row shows the predicted 3D facial landmarks.}
	\label{fig:3dfawDemo}
\end{figure*}

\begin{figure*}[t]
	\centering
	\includegraphics[width=0.99\textwidth]{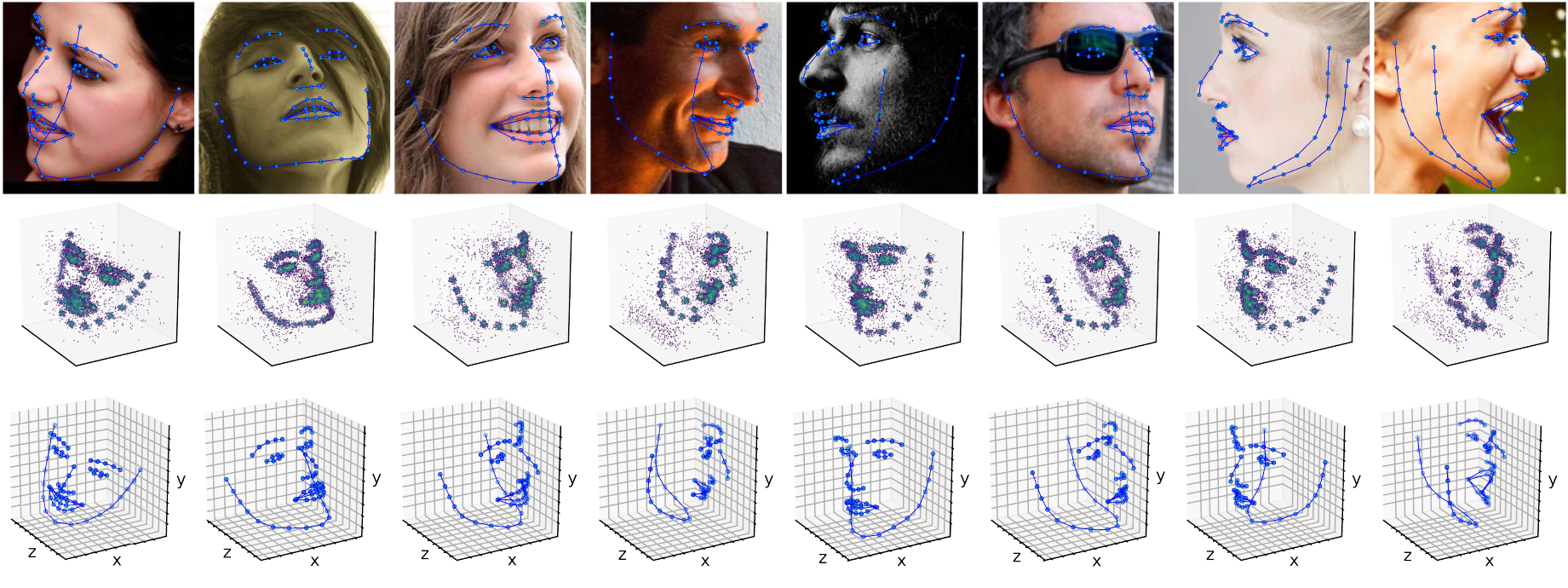}
	\caption{Example results of the proposed method on the AFLW2000-3D dataset. Images are arranged as the same as Fig.~\ref{fig:3dfawDemo}}
	\label{fig:aflwDemo}
\end{figure*}

\subsubsection{Evaluation on AFLW2000-3D}

We further evaluate our method on AFLW2000-3D to demonstrate the effectiveness of our method on face images with large pose and appearance variations.
Our method is trained on 300W-LP and tested on AFLW2000-3D.
Performance of both the 3D and 2D landmark localization is evaluated for thorough comparison.

Since there are a limited number of previous works on 3D facial landmark localization in the wild and the training code for most of them are not available, we consider the most recent state-of-the-art method FAN~\cite{Bulat2017HowFar}, with the code released by the authors, as the baseline for comparison.
Specifically, for 3D facial landmark localization, the baseline method FAN+Depth performs 2D landmark localization using FAN~\cite{Bulat2017HowFar} at first and then estimates the depth using ResNet~\cite{he2016deep}.
Comparisons of GTE and CED curves with the baseline method are shown in Table~\ref{tab:compAFLW3D} and Fig.~\ref{fig:compCED} respectively.
Benefiting from the end-to-end pipeline, the proposed method outperforms the strong baseline method considerably.

We also compare our method with other methods on 2D facial landmark localization.
In this case, only 2D coordinates are involved in the evaluation and the metric is the Normalized Mean Error (NME), where the normalized distance is square-root of the size of the bounding box enclosing all 2D landmarks.
Comparisons of NME across poses are reported in Table~\ref{tab:compAFLW2D}.
Note that the results of RCPR~\cite{burgos2013robust}, ESR~\cite{cao2014face}, and SDM~\cite{xiong2013supervised} are obtained from~\cite{zhu2016face} and these methods have been retrained on 300W-LP for adaptation to large poses.
As shown in Table~\ref{tab:compAFLW2D}, the proposed method achieve a superior performance especially for large poses.
Example results of our method on AFLW2000-3D are depicted in Fig.~\ref{fig:aflwDemo}.
It can be seen that our method is robust to occlusions and large appearance variations occurred in face images in the wild.

\section{Conclusion}
\label{Conclusion}
In this paper, we present the Joint Voxel and Coordinate Regression (JVCR) method for 3D facial landmark localization. 
First, we introduce the compact volumetric representation which encodes positions of all landmarks in a single volume.
In this way, the dimensionality of the representation could be reduced greatly compared with the conventional volumetric representation.
For robust and accurate 3D facial landmark localization, we perform the coarse-to-fine voxel regression and coordinate regression via the stacked hourglass network and 3D convolution network, respectively.
Hence, the joint voxel and coordinate regression could combine the merits of both heatmap regression based methods and coordinate regression based methods.
Moreover, the proposed method is able to produce satisfying results on face images with occlusions and large appearance variations. 
Experimental results on 3DFAW and AFLW2000-3D datasets demonstrate the superiority of our approaches in comparison with other state-of-the-art solutions.
In future work, we will investigate the proposed method further in the context of general 3D object landmark localization such as 3D human pose estimation in the wild.

\bibliographystyle{IEEEtran}
% argument is your BibTeX string definitions and bibliography database(s)
\bibliography{egbib}

% Generated by IEEEtran.bst, version: 1.13 (2008/09/30)
\begin{thebibliography}{10}
\providecommand{\url}[1]{#1}
\csname url@samestyle\endcsname
\providecommand{\newblock}{\relax}
\providecommand{\bibinfo}[2]{#2}
\providecommand{\BIBentrySTDinterwordspacing}{\spaceskip=0pt\relax}
\providecommand{\BIBentryALTinterwordstretchfactor}{4}
\providecommand{\BIBentryALTinterwordspacing}{\spaceskip=\fontdimen2\font plus
\BIBentryALTinterwordstretchfactor\fontdimen3\font minus
  \fontdimen4\font\relax}
\providecommand{\BIBforeignlanguage}[2]{{%
\expandafter\ifx\csname l@#1\endcsname\relax
\typeout{** WARNING: IEEEtran.bst: No hyphenation pattern has been}%
\typeout{** loaded for the language `#1'. Using the pattern for}%
\typeout{** the default language instead.}%
\else
\language=\csname l@#1\endcsname
\fi
#2}}
\providecommand{\BIBdecl}{\relax}
\BIBdecl

\bibitem{bulat2016convolutional}
A.~Bulat and G.~Tzimiropoulos, ``Convolutional aggregation of local evidence
  for large pose face alignment,'' in \emph{British Machine Vision Conference},
  2016, pp. 86.1--86.12.

\bibitem{Bulat2017HowFar}
------, ``How far are we from solving the 2d \& 3d face alignment problem? (and
  a dataset of 230,000 3d facial landmarks),'' in \emph{Proceedings of the IEEE
  International Conference on Computer Vision}, Oct 2017, pp. 1021--1030.

\bibitem{xiong2013supervised}
X.~Xiong and F.~De~la Torre, ``Supervised descent method and its applications
  to face alignment,'' in \emph{Proceedings of the IEEE Conference on Computer
  Vision and Pattern Recognition}, 2013, pp. 532--539.

\bibitem{cao2014face}
X.~Cao, Y.~Wei, F.~Wen, and J.~Sun, ``Face alignment by explicit shape
  regression,'' \emph{International Journal of Computer Vision}, vol. 107,
  no.~2, pp. 177--190, 2014.

\bibitem{ren2014face}
S.~Ren, X.~Cao, Y.~Wei, and J.~Sun, ``Face alignment at 3000 fps via regressing
  local binary features,'' in \emph{Proceedings of the IEEE Conference on
  Computer Vision and Pattern Recognition}, 2014, pp. 1685--1692.

\bibitem{bulat2016human}
A.~Bulat and G.~Tzimiropoulos, ``Human pose estimation via convolutional part
  heatmap regression,'' in \emph{European Conference on Computer Vision}.\hskip
  1em plus 0.5em minus 0.4em\relax Springer, 2016, pp. 717--732.

\bibitem{newell2016stacked}
A.~Newell, K.~Yang, and J.~Deng, ``Stacked hourglass networks for human pose
  estimation,'' in \emph{European Conference on Computer Vision}.\hskip 1em
  plus 0.5em minus 0.4em\relax Springer, 2016, pp. 483--499.

\bibitem{zhao2016fast}
R.~Zhao, Y.~Wang, C.~F. Benitez-Quiroz, Y.~Liu, and A.~M. Martinez, ``Fast and
  precise face alignment and 3d shape reconstruction from a single 2d image,''
  in \emph{European Conference on Computer Vision Workshops}.\hskip 1em plus
  0.5em minus 0.4em\relax Springer, 2016, pp. 590--603.

\bibitem{bulat2016two}
A.~Bulat and G.~Tzimiropoulos, ``Two-stage convolutional part heatmap
  regression for the 1st 3d face alignment in the wild (3dfaw) challenge,'' in
  \emph{European Conference on Computer Vision Workshops}.\hskip 1em plus 0.5em
  minus 0.4em\relax Springer, 2016, pp. 616--624.

\bibitem{gou2016shape}
C.~Gou, Y.~Wu, F.-Y. Wang, and Q.~Ji, ``Shape augmented regression for 3d face
  alignment,'' in \emph{European Conference on Computer Vision
  Workshops}.\hskip 1em plus 0.5em minus 0.4em\relax Springer, 2016, pp.
  604--615.

\bibitem{pavlakos2017coarse}
G.~Pavlakos, X.~Zhou, K.~G. Derpanis, and K.~Daniilidis, ``Coarse-to-fine
  volumetric prediction for single-image 3d human pose,'' in \emph{Proceedings
  of the IEEE Conference on Computer Vision and Pattern Recognition}.\hskip 1em
  plus 0.5em minus 0.4em\relax IEEE, 2017, pp. 1263--1272.

\bibitem{jeni2016first}
L.~A. Jeni, S.~Tulyakov, L.~Yin, N.~Sebe, and J.~F. Cohn, ``The first 3d face
  alignment in the wild (3dfaw) challenge,'' in \emph{European Conference on
  Computer Vision}.\hskip 1em plus 0.5em minus 0.4em\relax Springer, 2016, pp.
  511--520.

\bibitem{zhu2016face}
X.~Zhu, Z.~Lei, X.~Liu, H.~Shi, and S.~Z. Li, ``Face alignment across large
  poses: A 3d solution,'' in \emph{Proceedings of the IEEE Conference on
  Computer Vision and Pattern Recognition}, 2016, pp. 146--155.

\bibitem{cristinacce2006feature}
D.~Cristinacce and T.~F. Cootes, ``Feature detection and tracking with
  constrained local models.'' in \emph{British Machine Vision Conference},
  vol.~1, no.~2, 2006, p.~3.

\bibitem{saragih2011deformable}
J.~M. Saragih, S.~Lucey, and J.~F. Cohn, ``Deformable model fitting by
  regularized landmark mean-shift,'' \emph{International Journal of Computer
  Vision}, vol.~91, no.~2, pp. 200--215, 2011.

\bibitem{yu2013pose}
X.~Yu, J.~Huang, S.~Zhang, W.~Yan, and D.~N. Metaxas, ``Pose-free facial
  landmark fitting via optimized part mixtures and cascaded deformable shape
  model,'' in \emph{Proceedings of the IEEE International Conference on
  Computer Vision}, 2013, pp. 1944--1951.

\bibitem{cootes2001active}
T.~F. Cootes, G.~J. Edwards, and C.~J. Taylor, ``Active appearance models,''
  \emph{IEEE Transactions on Pattern Analysis and Machine Intelligence}, no.~6,
  pp. 681--685, 2001.

\bibitem{matthews2004active}
I.~Matthews and S.~Baker, ``Active appearance models revisited,''
  \emph{International Journal of Computer Vision}, vol.~60, no.~2, pp.
  135--164, 2004.

\bibitem{yang2017ed}
J.~Yang, Q.~Liu, and K.~Zhang, ``Stacked hourglass network for robust facial
  landmark localisation,'' in \emph{Proceedings of the IEEE Conference on
  Computer Vision and Pattern Recognition Workshops}.\hskip 1em plus 0.5em
  minus 0.4em\relax IEEE, 2017, pp. 2025--2033.

\bibitem{chu2017multi}
X.~Chu, W.~Yang, W.~Ouyang, C.~Ma, A.~L. Yuille, and X.~Wang, ``Multi-context
  attention for human pose estimation,'' in \emph{Proceedings of the IEEE
  Conference on Computer Vision and Pattern Recognition}, 2017, pp. 5669--5678.

\bibitem{sun2017integral}
X.~Sun, B.~Xiao, S.~Liang, and Y.~Wei, ``Integral human pose regression,''
  \emph{arXiv preprint arXiv:1711.08229}, 2017.

\bibitem{Tulyakov2017Viewpoint}
S.~Tulyakov, L.~A. Jeni, J.~F. Cohn, and N.~Sebe, ``Viewpoint-consistent 3d
  face alignment,'' \emph{IEEE Transactions on Pattern Analysis and Machine
  Intelligence}, 2017.

\bibitem{tompson2014joint}
J.~J. Tompson, A.~Jain, Y.~LeCun, and C.~Bregler, ``Joint training of a
  convolutional network and a graphical model for human pose estimation,'' in
  \emph{Advances in neural information processing systems}, 2014, pp.
  1799--1807.

\bibitem{wei2016convolutional}
S.-E. Wei, V.~Ramakrishna, T.~Kanade, and Y.~Sheikh, ``Convolutional pose
  machines,'' in \emph{Proceedings of the IEEE Conference on Computer Vision
  and Pattern Recognition}, 2016, pp. 4724--4732.

\bibitem{wu20153d}
Z.~Wu, S.~Song, A.~Khosla, F.~Yu, L.~Zhang, X.~Tang, and J.~Xiao, ``3d
  shapenets: A deep representation for volumetric shapes,'' in
  \emph{Proceedings of the IEEE Conference on Computer Vision and Pattern
  Recognition}, 2015, pp. 1912--1920.

\bibitem{ge20173d}
L.~Ge, H.~Liang, J.~Yuan, and D.~Thalmann, ``3d convolutional neural networks
  for efficient and robust hand pose estimation from single depth images,'' in
  \emph{Proceedings of the IEEE Conference on Computer Vision and Pattern
  Recognition}, vol.~1, 2017, p.~5.

\bibitem{ji20133d}
S.~Ji, W.~Xu, M.~Yang, and K.~Yu, ``3d convolutional neural networks for human
  action recognition,'' \emph{IEEE Transactions on Pattern Analysis and Machine
  Intelligence}, vol.~35, no.~1, pp. 221--231, 2013.

\bibitem{tieleman2012lecture}
T.~Tieleman and G.~Hinton, ``Lecture 6.5-rmsprop: Divide the gradient by a
  running average of its recent magnitude,'' \emph{COURSERA: Neural Networks
  for Machine Learning}, vol.~4, no.~2, pp. 26--31, 2012.

\bibitem{yin2008high}
L.~Yin, X.~Chen, Y.~Sun, T.~Worm, and M.~Reale, ``A high-resolution 3d dynamic
  facial expression database,'' in \emph{Automatic Face \& Gesture Recognition,
  2008. FG'08. 8th IEEE International Conference on}.\hskip 1em plus 0.5em
  minus 0.4em\relax IEEE, 2008, pp. 1--6.

\bibitem{zhang2014bp4d}
X.~Zhang, L.~Yin, J.~F. Cohn, S.~Canavan, M.~Reale, A.~Horowitz, P.~Liu, and
  J.~M. Girard, ``Bp4d-spontaneous: a high-resolution spontaneous 3d dynamic
  facial expression database,'' \emph{Image and Vision Computing}, vol.~32,
  no.~10, pp. 692--706, 2014.

\bibitem{gross2010multi}
R.~Gross, I.~Matthews, J.~Cohn, T.~Kanade, and S.~Baker, ``Multi-pie,''
  \emph{Image and Vision Computing}, vol.~28, no.~5, pp. 807--813, 2010.

\bibitem{jeni2017dense}
L.~A. Jeni, J.~F. Cohn, and T.~Kanade, ``Dense 3d face alignment from 2d video
  for real-time use,'' \emph{Image and Vision Computing}, vol.~58, pp. 13--24,
  2017.

\bibitem{sagonas2013300}
C.~Sagonas, G.~Tzimiropoulos, S.~Zafeiriou, and M.~Pantic, ``300 faces
  in-the-wild challenge: The first facial landmark localization challenge,'' in
  \emph{Proceedings of the IEEE International Conference on Computer Vision
  Workshops}, 2013, pp. 397--403.

\bibitem{kostinger2011annotated}
M.~K{\"o}stinger, P.~Wohlhart, P.~M. Roth, and H.~Bischof, ``Annotated facial
  landmarks in the wild: A large-scale, real-world database for facial landmark
  localization,'' in \emph{Proceedings of the IEEE International Conference on
  Computer Vision Workshops}.\hskip 1em plus 0.5em minus 0.4em\relax IEEE,
  2011, pp. 2144--2151.

\bibitem{Bhagavatula2017Faster}
C.~Bhagavatula, C.~Zhu, K.~Luu, and M.~Savvides, ``Faster than real-time facial
  alignment: A 3d spatial transformer network approach in unconstrained
  poses,'' in \emph{Proceedings of the IEEE International Conference on
  Computer Vision}, Oct 2017, pp. 4000--4009.

\bibitem{de20163d}
F.~H. de~Bittencourt~Zavan, A.~C. Nascimento, L.~P. e~Silva, O.~R. Bellon, and
  L.~Silva, ``3d face alignment in the wild: A landmark-free, nose-based
  approach,'' in \emph{European Conference on Computer Vision Workshops}.\hskip
  1em plus 0.5em minus 0.4em\relax Springer, 2016, pp. 581--589.

\bibitem{burgos2013robust}
X.~P. Burgos-Artizzu, P.~Perona, and P.~Doll{\'a}r, ``Robust face landmark
  estimation under occlusion,'' in \emph{Proceedings of the IEEE International
  Conference on Computer Vision}, 2013, pp. 1513--1520.

\bibitem{he2016deep}
K.~He, X.~Zhang, S.~Ren, and J.~Sun, ``Deep residual learning for image
  recognition,'' in \emph{Proceedings of the IEEE Conference on Computer Vision
  and Pattern Recognition}, 2016, pp. 770--778.

\end{thebibliography}

% that's all folks
\end{document}